\documentclass[10pt,twocolumn,letterpaper]{article}
\usepackage{todonotes}
\usepackage{multirow}
\usepackage{pifont}
\usepackage{colortbl}
\usepackage{comment}
\usepackage{makecell}

\usepackage{booktabs}
\usepackage{siunitx}

\usepackage[pagenumbers]{cvpr} 

\definecolor{cvprblue}{rgb}{0.21,0.49,0.74}
\usepackage[pagebackref,breaklinks,colorlinks,allcolors=cvprblue]{hyperref}

\def\paperID{221} 
\def\confName{CVPR}
\def\confYear{2026}

\title{edgeVLM: Cloud-edge Collaborative Real-time VLM based on Context Transfer}

\author{Chen Qian\\
Tsinghua University
\and
Xinran Yu\\
Tsinghua University
\and
Zewen Huang\\
Tsinghua University
\and
Danyang Li\\
Tsinghua University
\and
Qiang Ma\\
Tsinghua University
\and
Fan Dang\\
Beijing Jiaotong University
\and
Xuan Ding\\
Tsinghua University
\and
Guangyong Shang\\
Inspur Yunzhou Industrial Internet Co., Ltd
\and
Zheng Yang\\
Tsinghua University
}

\begin{document}
\maketitle
\begin{abstract}
Vision–Language Models (VLMs) are increasingly deployed in real-time applications such as autonomous driving and human–computer interaction, which demand fast and reliable responses based on accurate perception.
To meet these requirements, existing systems commonly employ cloud–edge collaborative architectures, such as partitioned Large Vision-Language Models (LVLMs) or task offloading strategies between Large and Small Vision-Language Models (SVLMs).
However, these methods fail to accommodate cloud latency fluctuations and overlook the full potential of delayed but accurate LVLM responses.
In this work, we propose a novel cloud–edge collaborative paradigm for VLMs, termed \textbf{Context Transfer}, which treats the delayed outputs of LVLMs as historical context to provide real-time guidance for SVLMs inference.
Based on this paradigm, we design \textbf{edgeVLM}, which incorporates both context replacement and visual focus modules to refine historical textual input and enhance visual grounding consistency.
Extensive experiments on three real-time vision-language reasoning tasks across four datasets demonstrate the effectiveness of the proposed framework. The new paradigm lays the groundwork for more effective and latency-aware collaboration strategies in future VLM systems.
Code will be publicly released before publication.
\end{abstract}    
\section{Introduction}
\label{sec:intro}

\begin{figure}[t]
  \centering
   \includegraphics[width=1\linewidth]{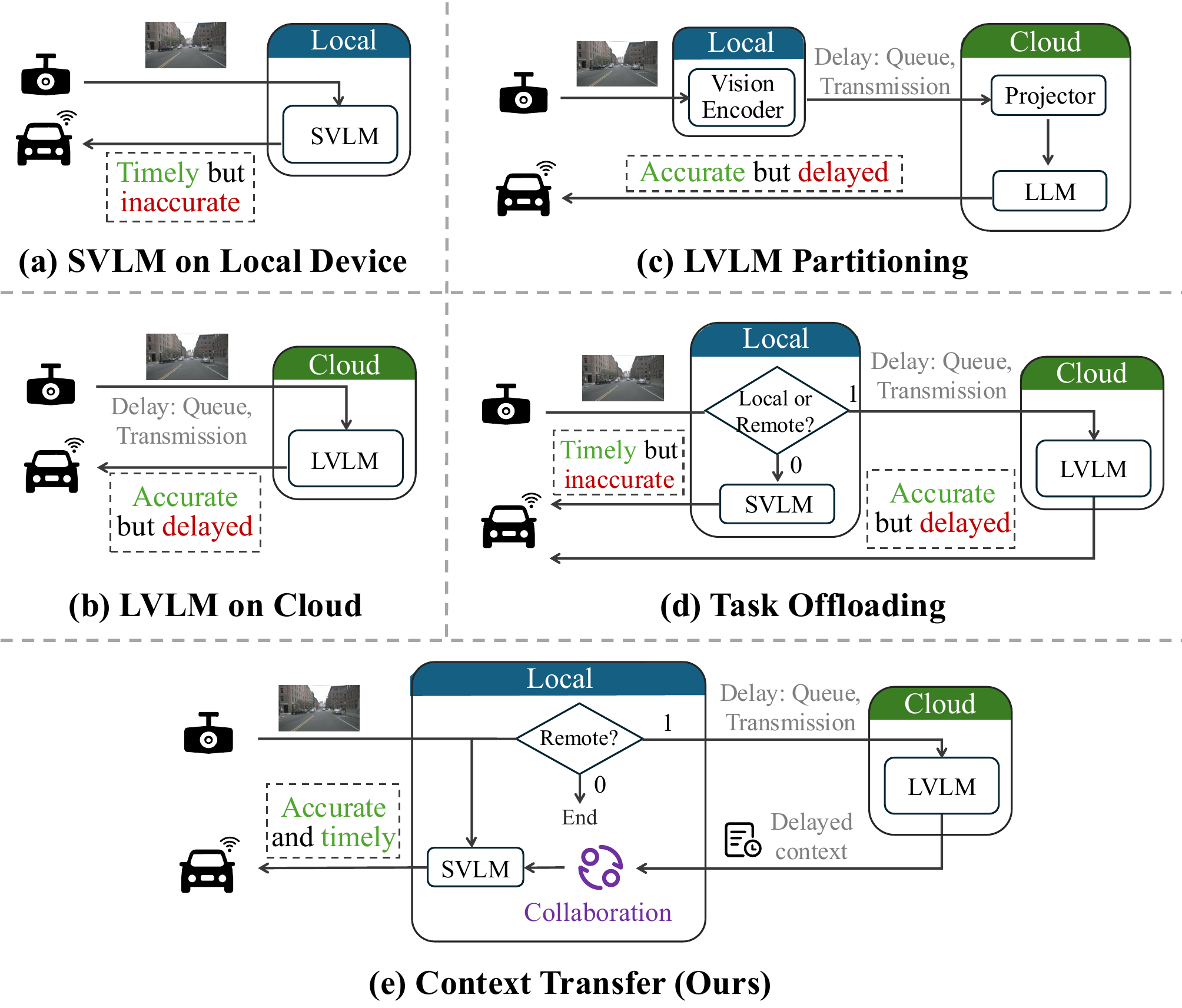}
   \caption{Overview of representative deployment and inference strategies for VLMs.}
   \label{fig:intro_allarchi}
\end{figure}

VLMs have witnessed rapid advancements in recent years. These models typically consist of a visual encoder, a language decoder, and a cross-modal fusion module, enabling them to perform a wide range of tasks such as image captioning~\cite{xue2025progress}, visual question answering~\cite{Fang_2025_CVPR}, and robotic perception~\cite{zhou2025physvlm,liu2025reasongrounder}. Notably, LVLMs~\cite{liu2025making}, empowered by large-scale model capacity and extensive training data, demonstrate superior generalization ability and reasoning performance across a broad range of downstream tasks. However, due to their substantial computational demands, these models are often deployed on the cloud servers, as illustrated in Fig.\ref{fig:intro_allarchi}(b). This deployment inevitably involve network transmission, model inference, and request queuing delays, posing challenges for scenarios that require real-time responses within seconds.

To overcome these limitations, recent studies have proposed lightweight SVLMs for edge deployment~\cite{yao2024minicpm, rang2025eve, lu2025bluelm}, enabling low-latency~\cite{vasu2025fastvlm}, on-device inference suitable for real-time scenarios like autonomous driving~\cite{lu2025can, sima2024drivelm, tian2024drivevlm}, augmented reality~\cite{pan2025metaspatial}, and smart wearable devices~\cite{huang2024vinci}, as illustrated in Fig.\ref{fig:intro_allarchi}(a). However, due to their limited model capacity, SVLMs still exhibit significantly lower inference accuracy than LVLMs~\cite{zhao2025stitch}.

To enable efficient and accurate real-time vision-language reasoning, recent studies~\cite{hu2024cloud, li2025distributed, hu2024laecips} have explored cloud–edge collaborative architectures. Existing approaches fall into two categories: (i) Model partitioning , which splits the LVLM across the cloud and edge based on hardware constraints. A representative example is Distributed VLM~\cite{li2025distributed}, depicted in Fig.\ref{fig:intro_allarchi}(c), which deploys the visual encoder on the edge and transmits extracted features to the cloud. (ii) Dynamic task offloading, which routes inputs based on task complexity and system conditions, as illustrated in Fig.\ref{fig:intro_allarchi}(d). For instance, ADAS~\cite{hu2024cloud} offloads tasks to the cloud for LVLM inference  by considering factors such as latency and quality of service (QoS). However, these approaches inherently depend on obtaining timely feedback from the LVLM to produce real-time outputs for the current frame, and they largely ignore the additional latency caused by network jitter and request queuing in cloud services.


To further explore more effective collaboration paradigms between large and small VLMs, we analyze the underlying factors that contribute to their performance gap and derive the following insights:

\noindent $\bullet$ \textbf{Utilization of High-Quality Textual Context.} Historical context plays a crucial role in generating high-quality outputs. Qwen-VL~\cite{bai2023qwen}, as a representative VLM, adopts an accumulative input structure in multi-turn interactions: each round of input consists of the input and response from the previous round, along with the current query. This context is progressively extended until the window reaches the maximum memory or model-defined limit. Such accumulation allows the model to integrate information across turns, enhancing semantic consistency and reasoning depth. Due to its limited generative capacity, SVLM often produces low-quality or even erroneous historical content, which propagates to subsequent steps and degrades its reasoning performance. Enhancing the quality of contextual history thus emerges as a key factor in improving the reasoning ability of small models.

\noindent $\bullet$ \textbf{Grounding of Semantically Salient Image Regions.} The ability to align visual content with textual descriptions is critical in vision-language reasoning~\cite{zong2025ground}. However, SVLMs often struggle in this aspect, particularly when dealing with fine-grained semantics or complex scenes. Prior studies~\cite{zheng2025deepeyes, xu2025visual, zhang2025chain} have demonstrated that LVLMs, leveraging their grounding capabilities, can automatically crop semantically rich image regions as refined inputs, thereby enhancing the focus and semantic interpretability in visual reasoning. Guiding small models to focus on task-relevant image regions enhances their ability to interpret complex visual and linguistic cues, ultimately improving their reasoning performance.

\textbf{Our Work:} Motivated by the critical role of accurate and task-relevant context in model reasoning, we propose a key design principle in this work: \textbf{Context Transfer}. Instead of relying on the delayed outputs from LVLMs as real-time feedback, they are reused as contextual priors to enhance future reasoning in SVLMs, as illustrated in Fig.\ref{fig:intro_allarchi}(e).

Building on this principle, we introduce \textbf{edgeVLM}, which incorporates two collaborative modules that guide the SVLM along both the linguistic and visual dimensions to achieve more accurate multimodal reasoning.

Motivated by the strong impact of historical context quality on VLM inference, we design a \textbf{Context Replacement Module (CRM)}. Specifically,
It extracts key textual segments from high-quality LVLM responses, reorganizes them based on task context, and embeds them as contextual history to replace the original low-quality history. By exploiting inter-frame correlations among temporally adjacent frames, these refined historical cues offer precise contextual guidance for current-frame inference, implicitly steering the small model toward semantically relevant visual evidence that was previously overlooked.

To exploit the grounding capabilities of large models, we introduce the \textbf{Visual Focus Module (VFM)}, which identifies semantically salient regions from historical images based on LVLM predictions. These regions are then used to guide the SVLM: non-essential patches are masked to reduce visual tokens and improve efficiency, while attention is focused on informative areas to enhance grounding and alignment. To adapt visual focus dynamically, the edge device performs feature matching between historical and current images. Semantically aligned regions are assigned higher weights, allowing the SVLM to inherit and shift attention toward task-relevant areas in the current frame.


The contributions of this work are summarized as follows:
\begin{itemize}
    \item We propose Context Transfer, a cloud–edge collaborative paradigm for real-time VLM inference. Instead of relying on the final predictions from LVLMs, this paradigm treats their delayed outputs as contextual priors to guide SVLMs, enabling a new form of efficient large–small model collaboration.
    \item Based on this principle, we design edgeVLM with two training-free collaborative modules: the Context Replacement Module and the Visual Focus Module, which guide the SVLMs to reuse high-quality outputs from the LVLMs along the linguistic and visual dimensions, respectively.
    \item Extensive experiments on four public benchmarks validate the effectiveness of edgeVLM, showing consistent improvements over prior cloud–edge collaborative approaches.
\end{itemize}

\begin{figure*}[t]
  \centering
   \includegraphics[width=1\linewidth]{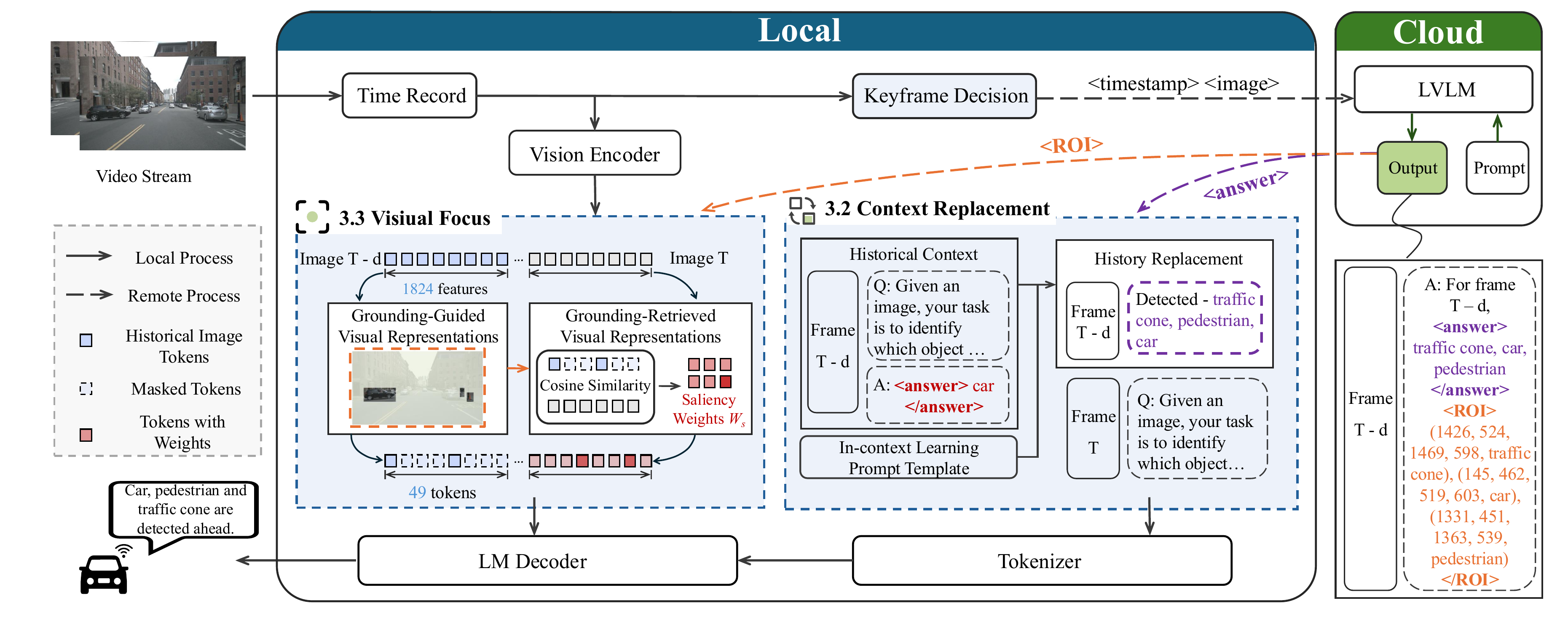}

   \caption{Overview of the proposed collaboration framework, \textbf{edgeVLM}. The system takes timestamped video streams as input and performs two parallel operations: uploading selected keyframes to the cloud-based LVLM for processing, and conducting local inference using the SVLM. Given potential cloud latency, delayed LVLM outputs are reused as historical context. These outputs guide the SVLM through two modules, Context Replacement and Visual Focus, to improve the quality of real-time predictions.}
   \label{fig:archi}
\end{figure*}

\section{Related Work}
\label{sec:related_work}
\subsection{Cloud-Edge Collaborative VLMs}

Driven by the growing demand for real-time VLM-based processing, applications such as autonomous driving~\cite{lu2025can} are shifting from modular to end-to-end pipelines~\cite{shao2024lmdrive, li2025generative}, where latency sensitivity is critical and excessive delays can lead to system failure.


To enable fast and robust responses, cloud-edge collaborative architectures for VLMs have attracted increasing attention. Existing collaboration paradigms can be categorized into two types. The first category is model partitioning~\cite{li2025distributed, zhang2025vavlm}. To improve processing efficiency, Distributed VLM~\cite{li2025distributed} offload the visual encoder to the edge device, which encodes input images and transmits features to the cloud for processing and decoding. VaVLM~\cite{zhang2025vavlm} reduces transmission overhead by uploading only the region-of-interest (ROI) of images. These parallel architecture entirely rely on stable and responsive cloud services, making it difficult to guarantee robust frame-wise inference in real-time settings. The second category involves dynamic task allocation to different models according to predefined offloading strategies. Some systems~\cite{hu2024cloud} formulate the offloading decision as an optimization problem, incorporating multiple system-level factors such as Quality of Service (QoS) and latency. Frameworks such as LAECIPS~\cite{hu2024laecips} pretrain a dedicated task classifier to estimate task complexity and offload difficult samples to cloud-based large models. While these approaches leverage the complementary strengths of large and small models to reduce cloud-side computation, they overlook the continued guidance that high-quality LVLM responses can provide for subsequent inference. Recognizing this underexplored potential, we propose a real-time collaborative framework that integrates delayed LVLM outputs into SVLM context to enable continual refinement and robust inference.


\subsection{Historical Context in VLMs}
Context refers to the sequence of prior interactions retained to guide subsequent inferences, essential for maintaining coherence in multi-turn tasks. In VLMs, this mechanism is adapted from Large Language Models (LLMs), which use fixed-length context windows to preserve relevant token sequences and enhance reasoning consistency. Prior research on LLMs has shown that adjusting context improves model performance, as reflected in translation accuracy~\cite{sung2024context} and dialogue consistency~\cite{wei2024hidden}. Similarly, VLMs leverage effective context preservation to significantly improve robotic navigation in embodied AI scenarios~\cite{habibpour2025history}, reducing the risk of decision paralysis and oscillatory behavior. Building on this advantage, the proposed edgeVLM retain the output from LVLMs as high-quality context to guide the SVLMs, helping it focus on salient content and improve inference quality.

\section{Architecture}
\label{sec:method}

Real-time vision-language reasoning tasks pose dual challenges in terms of processing speed and inference accuracy. To achieve robust and high-precision outputs, this work leverages the strengths of LVLMs in textual understanding and visual perception to assist SVLMs in real-time. This section illustrates the proposed collaborative architecture using a multi-object recognition task in autonomous driving as a representative example.

\subsection{Workflow}
\label{sec:workflow}
The front-facing camera continuously captures frames and sends them to the edge device. For each timestamped frame, the edge decides whether to upload it to the cloud. Each frame is also processed locally by the SVLM for real-time inference. For uploaded frames, the returned LVLM responses are handled according to latency: results within $\tau$ second are adopted and stored, while delayed outputs are retained for later use, and real-time results from the SVLM are adopted for the current frame.

As illustrated in Fig.\ref{fig:archi}, the output for frame $T-d$ returned by the LVLM contains two key components: (1) the response content to the given query, denoted as \texttt{<answer>}; and (2) the attended image regions during inference, indicated as \texttt{<ROI>}. The two outputs respectively provide textual and visual guidance for SVLM inference at frame $T$.


\subsection{Context Replacement}
\label{sec:context_rep}
In multi-turn vision-language reasoning tasks, VLMs typically take an image and a textual question from the user as the current input. To support continuous multimodal understanding, VLMs often retain part of the historical context during inference. We find that replacing the history generated by SVLM with accurate historical context supplied by a delayed LVLM guides the SVLM to produce more accurate predictions for the current frame. To theoretically justify this empirical observation, consider the following sequential prediction formulation.

Let the prediction at time $t$ be
\begin{equation}
\hat{\mathbf{y}}_{t}=f(\mathbf{x}_{t},\mathbf{h}_{t-1}),
\end{equation}
where $\mathbf{x}_{t}$ is the current input and $\mathbf{h}_{t-1}$ is the  historical state from frame $(t\!-\!1)$. Denote by $\mathbf{h}^{\star}_{t-1}$ the accurate history and by $\mathbf{h}^{\star}_{t-1}+\mathbf{e}$ an inexact history. A first-order expansion in the historical state gives
\begin{equation}
f(\mathbf{x}_{t},\mathbf{h}^{\star}_{t-1}+\mathbf{e})
\approx
f(\mathbf{x}_{t},\mathbf{h}^{\star}_{t-1}) + J_h\,\mathbf{e},
\label{eq:firstorder}
\end{equation}
\begin{equation}
J_h:=\frac{\partial f}{\partial \mathbf{h}}\Big|_{(\mathbf{x}_{t},\mathbf{h}^{\star}_{t-1})}.
\end{equation}
Hence the induced output deviation is bounded by
\begin{equation}
\big\| f(\mathbf{x}_{t},\mathbf{h}^{\star}_{t-1}+\mathbf{e})
      - f(\mathbf{x}_{t},\mathbf{h}^{\star}_{t-1}) \big\|
\le \|J_h\|\,\|\mathbf{e}\|.
\label{eq:lipschitz}
\end{equation}
Let $\ell = \big\|\cdot\big\|^{2}$, the expected excess loss at time $t$ satisfies
\begin{equation}
\mathbb{E}[\Delta \ell]
\;\lesssim\;
\mathrm{tr}\!\big(J_h\,\Sigma_{e}\,J_h^{\top}\big),
\end{equation}
where $\Sigma_{e}$ is the covariance matrix of the historical error. Therefore, a more accurate history (smaller $\Sigma_e$) yields a smaller expected loss at frame $t$. This insight informs the design of two novel methods for enabling textual-level collaboration across LVLMs and SVLMs.

As detailed in Sec.~\ref{sec:workflow}, the \texttt{<answer>} produced by the LVLM is continuously refreshed and preserved in the edge-side cache.
At frame \textit{T}, the system locates in the edge cache the latest LVLM result corresponding to a prior frame \textit{T - d}. If the elapsed time $d \leq \delta$, that LVLM result replaces the historical output produced by the SVLM for frame \textit{T - d} and is incorporated into the context used to condition inference on frame \textit{T}.

Since adjacent frames are highly correlated, higher-quality historical information more reliably steers the model toward activating relevant tokens and suppressing irrelevant ones.

Given the limited memory capacity of edge devices and the context window constraints of lightweight models, naively stacking multi-turn inputs may lead to token truncation. Moreover, we design an in-context learning~\cite{kang2025context} prompt template to compress and summarize dialogue history for the SVLM.

\textbf{Example}. In the autonomous driving setup, the edge-side SVLM predicts “car” on the historical frame \textit{T - d}. If the corresponding cloud response arrives within the staleness bound $d \leq \delta$, we replace the history with the corrected set from LVLM “traffic cone, car, pedestrian.” We then convert the Q\&A-style history into a compressed snippet: “This is the image from d seconds ago; detected traffic cone, car, pedestrian.” This snippet is injected into the current input as prior.

To summarize, with the textual collaboration strategy in place, the SVLM receives an input structure at frame T that corresponds to the right side of the Context Replacement module in Fig.~\ref{fig:archi}.
This input incorporates distilled historical answers from the LVLM, enabling the SVLM to perform accurate and efficient reasoning.

\subsection{Visual Focus}
\label{sec:visual_fo}

LVLMs, equipped with strong grounding capabilities, identifies task-relevant image regions and transmits them to the edge-side SVLMs. Based on this, SVLMs can proactively focus on critical visual regions from two perspectives.

\begin{figure}[t]
  \centering
   \includegraphics[width=0.7\linewidth]{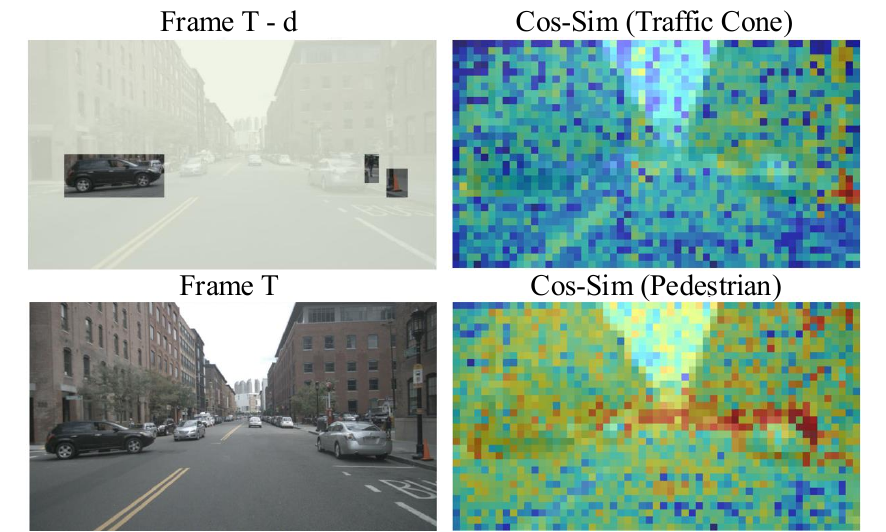}
   \caption{ROI-based region retrieving via cosine similarity. Given ROIs of traffic cone and pedestrian from frame \textit{T – d}, cosine similarity highlights related regions in frame \textit{T} to guide the attention of the SVLM.}
   \label{fig:cosine}
   \setlength{\belowcaptionskip}{-4pt}
\end{figure}

\subsubsection{Grounding-Guided Visual Representation} The ROIs returned by the LVLM guide the grounding of the SVLM on historical frames. Most VLMs, including the Qwen-2.5-VL-3B~\cite{Qwen2.5-VL}, use a Vision Transformer (ViT)~\cite{dosovitskiy2020image} that divides a resized input image into patches and encodes them into visual features (e.g., 1824 features for a 1600 × 900 image). However, many of these features are irrelevant to the reasoning task. To accelerate inference and enhance focus, ROIs are mapped to corresponding patches, and only task-relevant ones are selected. This selection is formally defined as:
\begin{equation}
\mathcal{P}_{ROI} = \left \{ p_i \in \mathcal{P} \mid p_i\cap ROI\ne \emptyset \right \},
  \label{eq:roi_patch}
\end{equation}
where $\mathcal{P}$ denotes the set of all image patches, and $p_i$ is a patch whose spatial overlap with the ROI is non-empty. In the case of multi-object recognition, the largest ROI is selected for each target category. For the case illustrated in Fig.~\ref{fig:archi}, the number of visual tokens selected after projection is reduced to 49, thereby reducing the computational cost.

\subsubsection{Grounding-Retrieved Visual Representation}The ROIs provided by the LVLM support cross-frame retrieving, guiding the SVLM to focus on relevant regions in subsequent frames. Given the temporal continuity, ROI features from frame \textit{T - d} can be matched to visual features in frame \textit{T}. Let $\mathcal{P}_{T-d}^{(c)}$ denotes the set of patch indices for category $c$ on frame \textit{T - d}, and $\mathbf{v}_{i}^{(T-d)}$ is the visual feature of the $i$-th patch. Thus, feature vector of object category $c$ on frame \textit{T-d} is:
\begin{equation}
\mathbf{f}^{(c)} = \frac{1}{|\mathcal{P}^{(c)}_{T-d}|} \sum_{i \in \mathcal{P}^{(c)}_{T-d}} \mathbf{v}^{(T-d)}_i.
  \label{eq:mean_f}
\end{equation}

We compute the cosine similarity between ROI features from frame \textit{T - d} and all visual features from frame \textit{T}, thereby obtaining similarity maps $S^{(c)}$ for each potential category $c$ in frame \textit{T}.
\begin{equation}
\mathbf{S}^{(c)} = \left[ \frac{ \mathbf{f}^{(c)} \cdot \mathbf{v}^{(T)}_j }{ \|\mathbf{f}^{(c)}\| \cdot \|\mathbf{v}^{(T)}_j\| } \right]_{j=1}^{N},
  \label{eq:sim_vec}
\end{equation}
where $v^{(T)}_j$ denotes the feature of the $j$-th patch on frame \textit{T}, and $N$ is the total number of patches. As shown in Fig.\ref{fig:cosine}, the LVLM accurately detects and localizes small objects such as traffic cones and pedestrians in frame \textit{T - d}, and their mapped similarity heatmaps on frame \textit{T} effectively indicate the potential locations after viewpoint shift, providing precise visual guidance for the SVLM. 
This cross-frame computation yields patch-level similarity scores, which can be aggregated into a unified similarity vector $s$, defined as:
\begin{equation}
s = \frac{1}{C} \sum_{c=1}^{C} S^{(c)}.
  \label{eq:sim_s}
\end{equation}
Then $s$ can be used to construct the grounding-retrieved saliency weight $W_s$, formulated as follows:
\begin{equation}
W_{s} = \mu \cdot \frac{1}{1 + e^{-s}} + b,
  \label{eq:attention_w}
\end{equation}
where \(\mu\) and \(b\) are tunable scaling parameters used to control the visual weights. Here, we enforce $0.5\mu + b = 1$ in our experiments.

To further adapt the visual representation based on semantic associations, we introduce a gated token update mechanism that enables dynamic modulation of visual tokens. The update is formulated as:
\begin{equation}
t\leftarrow \alpha \cdot t+(1 - \alpha) \cdot W_s \cdot t.
  \label{eq:token_update}
\end{equation}
Here, $t$ denotes the original visual token, and $\alpha \in [0,1]$ is a control coefficient that balances the weights between known object categories and newly emerging targets. The dynamically predicted weight $W_s$ modulates the visual tokens based on feature similarity, enabling adaptive token refinement under the guidance of semantic consistency.

\subsection{Keyframe Decision}

\begin{figure}[h]
  \centering
\setlength{\belowcaptionskip}{-3pt}
\setlength{\abovecaptionskip}{-1pt}
   \includegraphics[width=0.9\linewidth]{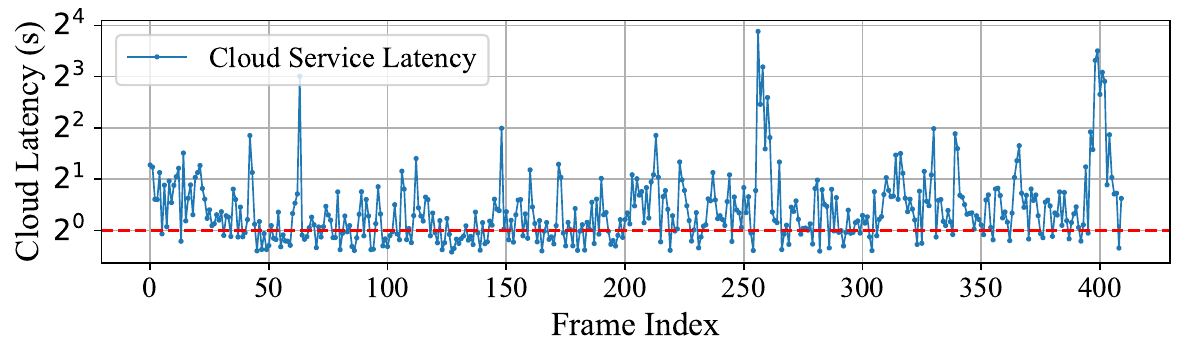}
   \caption{Experimental latency evaluation of cloud services over a 5G network.}
   \label{fig:cloud_delay}
\end{figure}

Under a 5G network environment, we invoke the Qwen-VL-2.5-72B API hosted on Alibaba Cloud and record the per-frame response latency, as illustrated in the Fig.\ref{fig:cloud_delay}. The results show that adjacent frames exhibit similar latency patterns. To enhance the efficiency of LVLM invocations and dynamically assess network conditions, the system continuously monitors how many preceding frames have not yet received responses. Based on this observation, we introduce a hyperparameter $K_{delay}$, which serves as a threshold: when the number of unreturned frames accumulates to $K_{delay}$, the system infers that the network condition has deteriorated and skips uploading the current frame.

We further regulate LVLM call frequency with a hyperparameter $K_{call}$. For frame $T$, if the most recent LVLM result is from frame $T - i$ with $i \le K_{call}$, that history is considered sufficiently fresh to guide the SVLM, and frame $T$ is not uploaded. This rule takes precedence over $K_{delay}$.

\vspace{-3pt}

\section{Experiments}
\label{sec:experiments}

\subsection{Experiments Setting}
\subsubsection{Datasets} We evaluate the proposed architecture on three real-time tasks across four datasets.

\noindent\textbf{Real-time multi-object recognition.} (i) nuScenes~\cite{nuscenes2019}: We sample images at 1 FPS from this autonomous driving dataset, which finally includes 156 scenes (avg. 20s each) with 10 object categories. (ii) BDD100K (MOT)~\cite{yu2020bdd100k}: We sample the driving video frame at 1 FPS, which includes 200 scenes (avg. 40s each) with 10 object categories.

\noindent\textbf{Real-time gesture recognition.} IPN Hands~\cite{bega2020IPNhand}: A gesture recognition dataset with 14 gesture classes, sampled at 2 FPS from 52 close-range videos (avg. 120s each) to capture dynamic motion.

\noindent\textbf{Real-time video frame caption.} Actions for Cooking Eggs (ACE)~\cite{shimada2012kitchen}: A kitchen scene dataset with 8 cooking actions, sampled at 1 FPS from 10 videos (avg. 194s each).

\subsubsection{Metrics}
We adopt commonly used evaluation metrics~\cite{durand2019learning, nam2017maximizing, huang2017multi} for multi-object recognition task, which are Micro-F1, Macro-F1 and 0-1 Exact Match (0-1 EM).
0-1 EM equals 1 only if the predicted result exactly matches the ground truth label.







We model both gesture recognition and video frame captioning as frame-level single-label classification tasks, and use accuracy (Acc) as the evaluation metric.

Note that the evaluated tasks operate under real-time constraints (1–2 FPS). Following DriveVLM \cite{tian2024drivevlm}, we set $\tau = 1$ in experiments.

\subsubsection{Implementation Details}
Experiments are conducted over a commercial 5G network. The average uplink bandwidth was 211 Mbps, downlink 72 Mbps, with measured round-trip latency around 42 ms. To remove variability from network jitter and queuing, we execute the cloud-side LVLM on every frame and log its streaming outputs and the per-frame latency, as shown in Fig.~\ref{fig:cloud_delay}. All collaboration strategies are evaluated against these identical per-frame traces for a fair comparison.

To balance processing speed and model performance, we retain two previous frames as context. All experiments on the edge side are conducted using a GeForce RTX 4090 GPU with 24 GB of memory.

The following abbreviations are used in this section: Grounding-Guided Visual Representation (GGVR), and Grounding-Retrieved Visual Representation (GRVR).


\vspace*{-2pt}
\subsection{Hyperparameter Study}
\begin{figure}[t]
  \centering
   \includegraphics[width=1\linewidth]{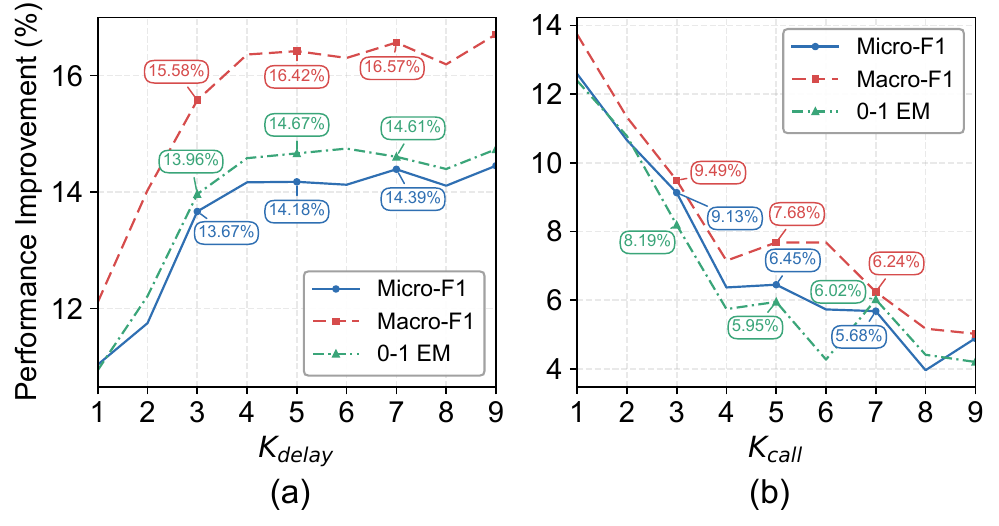}
   \caption{Absolute performance gain of edgeVLM over SVLM across different settings. (a) shows the impact of $K_{delay}$ with $K_{call} = 0$. (b) represents the impact of $K_{call}$ with $K_{delay} = 4$.}
   \label{fig:exp_k}
\end{figure}

\textbf{LVLM Call.} To balance the system performance and the cost of invoking the LVLM, we introduce two hyperparameters, $K_{delay}$ and $K_{call}$, in the Keyframe Decision module. A larger $K_{delay}$ leads to more frequent LVLM invocations. We conduct evaluations on the nuScenes dataset, and the resulting system performance is shown in Fig.\ref{fig:exp_k}(a). $K_{delay} = 4$ is selected as a trade-off choice. Under the 5G network, we recorded the LVLM utilization under different $K_{call}$ settings, as shown in Tab.\ref{tab:K_call}. \textbf{LCR (LVLM Call Ratio)} is the fraction of inference steps that trigger the cloud LVLM. \textbf{LHL (LVLM History Lag)} is the average frame offset $i$ between the current frame $t$ and the reused LVLM reply from frame $t - i$, computed over all frames that actually consume LVLM history. When $K_{call}$ increases from 2 to 3, the LVLM invocation frequency drops sharply, while the overall system performance remains relatively stable as shown in Fig.\ref{fig:exp_k}(b).

\noindent\textbf{Visual Guidance.} In the GRVR module, $\alpha$ and $\mu$ are introduced to regulate the visual guidance. Considering multi-metric performance and robustness, we adopt $(\alpha, \mu) = (0.7, 1.0)$.

\vspace*{-4pt}
\begin{table}[t]
\centering
\renewcommand{\arraystretch}{1.3}
\setlength{\tabcolsep}{3pt} 
\small{
\begin{tabular}{c|ccccccccc}
\toprule
$K_{call}$ & 1 & 2 & 3 & 4 & 5 & 6 & 7 & 8 & 9 \\
\hline
LCR(\%) & 69.1 & 38.9 & 27.3 & 23.5 & 19.0 & 18.4 & 17.8 & 14.4& 13.0\\
LHL(s) & 1.89 & 2.56 & 3.31 & 3.79 & 4.12 & 4.54 & 5.14 & 5.48 & 5.66 \\

\bottomrule
\end{tabular}
}
\caption{Impact of $K_{call}$ on LVLM utilization.}
\label{tab:K_call}
\vspace{-10pt}
\end{table}

\begin{table*}[t]
\centering
\setlength{\tabcolsep}{4pt}  
\renewcommand{\arraystretch}{1.2}  
\small{
\begin{tabular}{ccccccccccccc}
\toprule
\multirow{2}{*}{\textbf{Method}}
& \multicolumn{4}{c}{\textbf{nuScenes}}
& \multicolumn{4}{c}{\textbf{BDD100K MOT}}
& \multicolumn{2}{c}{\textbf{IPN Hand}}
& \multicolumn{2}{c}{\textbf{ACE}} \\
\cmidrule(lr){2-5} \cmidrule(lr){6-9} \cmidrule(lr){10-11} \cmidrule(lr){12-13}
& Micro-F1 & Macro-F1 & 0-1 EM &LCR
& Micro-F1 & Macro-F1 & 0-1 EM & LCR
& Acc &LCR
& Acc &LCR\\
\hline
SVLM Only (Edge) & 55.36 & 38.51 & 13.80 &0 & 61.56 & 30.79 & 6.59 &0 & 36.68 &0  & 22.86 &0  \\
LVLM Only (Cloud) & 43.84 & 33.81 & 18.98 &100  & 49.79 & 24.59 & 21.77 &100 & 22.55 &100 & 60.14 &100\\
Distributed VLM~\cite{li2025distributed} & 43.35 & 33.63 & 18.84 &100 & 48.28 & 23.55 & 20.89 &100 & 21.47 &100 & 59.43 &100 \\
\hline
\multicolumn{13}{c}{\textbf{LVLM Call Ratio (20-30\%)}}\\
\hline
LAECIPS~\cite{hu2024laecips} & 58.06 & 39.01 & 17.04 &29.3 & 61.70 & 30.48 & 12.88 & 28.7 & 38.86 & 28.8& 32.57 &28.0\\
ADAS~\cite{hu2024cloud} & 56.43 & 40.28 & 18.43 &28.2& 62.10 & 31.54 & 13.61 & 29.5 & 37.09 &25.6 & 33.29 & 28.5\\
\rowcolor{blue!10}
\textbf{edgeVLM} & \textbf{64.49} & \textbf{48.00} & \textbf{21.99} & \textbf{27.3} & \textbf{68.44} & \textbf{34.76} & \textbf{23.66}& \textbf{28.4} & \textbf{46.20} & \textbf{20.7} &\textbf{34.71} & \textbf{25.9} \\
\hline
\multicolumn{13}{c}{\textbf{LVLM Call Ratio (35-45\%)}}\\
\hline
LAECIPS & 57.24 & 39.39 & 19.58 & 44.5 & 61.88 & 30.90 & 15.98 &44.6 & 37.23 &44.0&\textbf{40.29} & 44.7\\
ADAS & 57.52 & 41.13 & 20.03 & 43.7& 62.55 & 30.62 & 16.88& 44.2 & 35.87 & 43.8 &39.71 & 43.6\\
\rowcolor{blue!10}
\textbf{edgeVLM} & \textbf{66.22} & \textbf{50.04} & \textbf{24.51} & \textbf{43.2}& \textbf{70.63} & \textbf{36.41} & \textbf{26.88}& \textbf{43.7} & \textbf{47.42} & \textbf{35.4}&38.86 &\textbf{35.1} \\
\hline
\multicolumn{13}{c}{\textbf{LVLM Call Ratio (50-60\%)}}\\
\hline
LAECIPS & 56.28 & 39.06 & 21.18 & 55.3 & 61.91 & 30.28 & 17.56 & 59.2 & 34.51 & 58.7& 45.71 & 54.0\\
ADAS & 58.03 & 42.23 & 22.13 &55.9& 62.88 & 30.62 & 19.99 & 57.4 & 35.33 & 56.2& 43.23 &54.8\\
\rowcolor{blue!10}
\textbf{edgeVLM} & \textbf{67.24} & \textbf{51.35} & \textbf{25.00} & \textbf{52.8} & \textbf{71.70} & \textbf{37.31} & \textbf{29.16}& \textbf{57.3} & \textbf{49.73} & \textbf{55.7}&\textbf{46.14} & \textbf{52.9} \\
\hline
\multicolumn{13}{c}{\textbf{LVLM Call Ratio (100\%)}}\\
\hline
\textbf{edgeVLM}& 71.15 & 57.11 & 29.41 & 100 & 74.12& 39.66& 35.08& 100 & 52.45&100 & 63.00&100 \\
\bottomrule
\end{tabular}
}
\caption{Performance comparison of different architectures across four datasets (values in \%). All methods adopt Qwen-VL-2.5-3B as the SVLM and Qwen-VL-2.5-72B as the LVLM.}
\label{tab:overall_table}
\end{table*}

\begin{figure*}[t]
  \centering
   \includegraphics[width=1\linewidth]{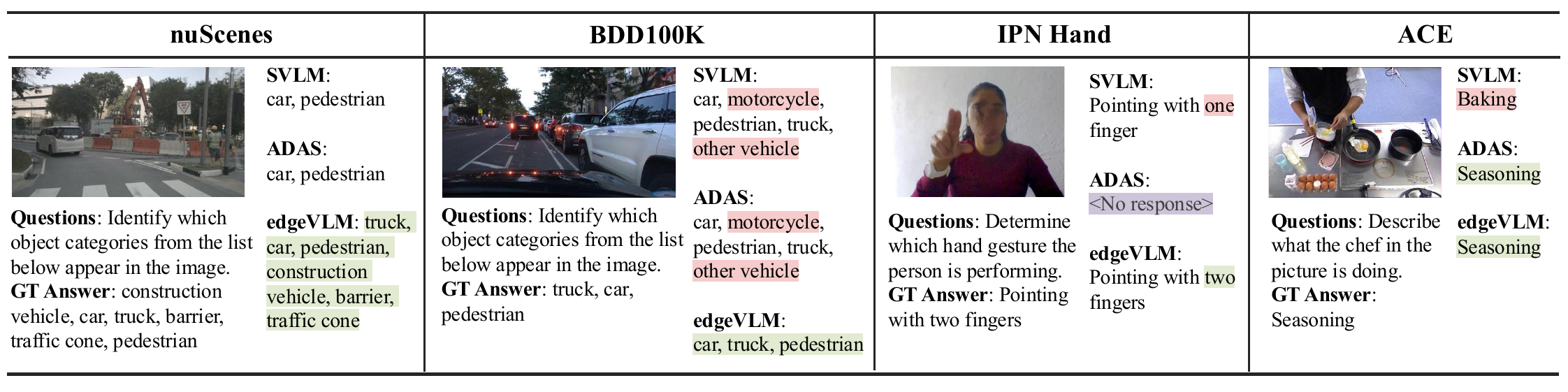}
   \caption{Qualitative Results. Illustration of input questions, ground-truth answers, and system predictions from four datasets.}
   \label{fig:overall_fig}
   \vspace{-4pt}
\end{figure*}

\subsection{System-level Evaluation}
\subsubsection{Overall Performance}
To evaluate the effectiveness of different architectures, we assess six methods: edge-based Qwen2.5‑VL‑3B, cloud-based Qwen2.5‑VL‑72B~\cite{Qwen2.5-VL}, Distributed VLM, LAECIPS, ADAS, and edgeVLM. The results are presented in Tab.\ref{tab:overall_table}.


For the fourth task, single-image inputs and short outputs allow rapid LVLM feedback, making full LVLM invocation clearly advantageous. For the other tasks, due to cloud latency, methods that rely entirely on LVLM yield lower F1 scores than collaborative approaches. However, the 0–1 EM of LVLM Only remains higher than that of SVLM Only, showing its superior accuracy when responses are timely. Additionally, Qwen2.5-VL-3B produces visual features larger than the original image, making Distributed VLM less effective than directly invoking the LVLM.

\begin{figure*}[t]
  \centering
  \setlength{\belowcaptionskip}{-2pt}
   \includegraphics[width=1\linewidth]{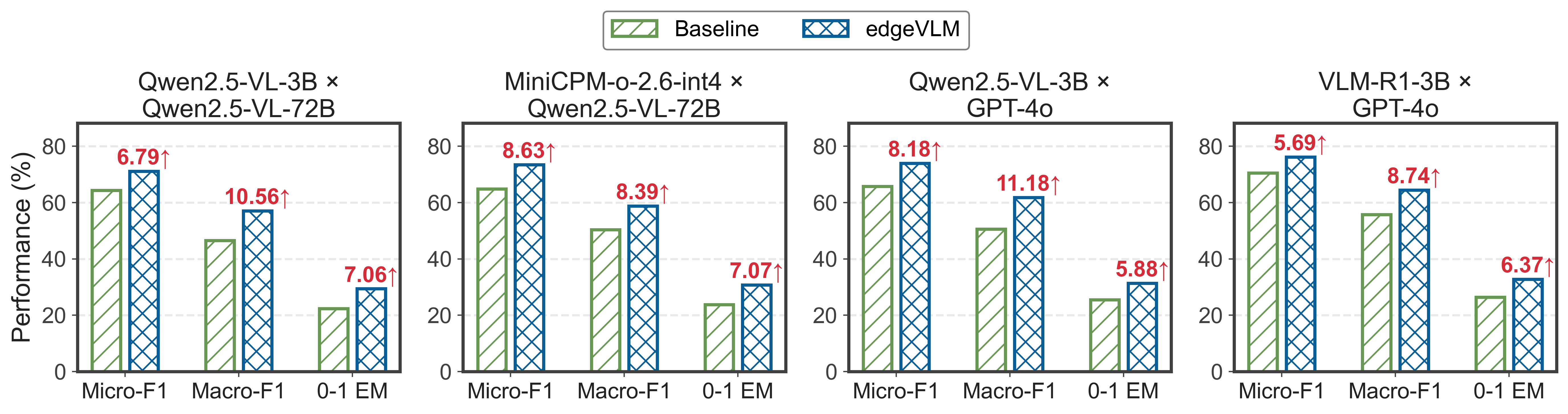}
   \caption{Adaptability Study. Testing the edgeVLM architecture with various SVLM–LVLM combinations on nuScenes. VLM-R1 is a small model further enhanced by reinforcement learning and knowledge distillation.}
   \label{fig:svlm_generalization}
\end{figure*}



By tuning the hyperparameters, we evaluate the system performance of edgeVLM under different LCR settings and compare it with alternative architectures. ADAS adopts latency-adaptive offloading strategies, where LVLM invocation depends on the current cloud service state. As a result, it tends to use a relatively low LCR on the first three datasets under realistic deployment. In contrast, LAECIPS triggers LVLM calls based on task difficulty and thus chooses a higher LCR on these datasets. Overall, however, both schemes deliver lower practical performance than edgeVLM, which more effectively exploits historical information even with a comparatively low LCR.

To provide a more intuitive comparison, we select four representative examples for qualitative analysis. As shown in Fig.\ref{fig:overall_fig}, for the real-time multi-object recognition task, all three methods rely on the SVLM, with edgeVLM achieving the best performance due to guidance from the historical results of the LVLM. In the gesture recognition task, as two images need to be transmitted and processed simultaneously, the LVLM fails to return valid results in real-time, leading to \textit{No response} for ADAS. For the video-frame captioning task, both ADAS and edgeVLM adopt the results quickly returned by the LVLM.

\vspace*{-4pt}
\subsubsection{Latency Analysis}

\begin{table}[t]
\centering
\renewcommand{\arraystretch}{1}
\setlength{\tabcolsep}{11pt} 
\small
\begin{tabular}{c c c}
\toprule
\textbf{System} & \textbf{Stage} & {\textbf{Time (ms)}} \\
\midrule
LVLM Only
  & \textbf{Total}      & \textbf{2110.83} \\
\midrule
SVLM Only
  & \textbf{Total}      & \textbf{581.71} \\
\midrule
\multirow{4}{*}{edgeVLM}
  & CRM-induced Overhead & 19.93 \\
  & VFM-induced Overhead        & 31.27 \\
  & Inference          & 511.20 \\
\rowcolor{blue!10}
  & \textbf{Total}      & \textbf{562.40} \\
\bottomrule
\end{tabular}
\caption{Execution time breakdown by system and stage (ms).}
\label{tab:exp_time_analysis}
\vspace{-2pt}
\end{table}

To assess module-induced computational overhead, we report the average execution time of different systems in Tab.~\ref{tab:exp_time_analysis}. CRM and VFM introduce additional processing latency. However, by reducing the number of visual tokens, VFM lowers the compute over the entire inference pipeline and offsets part of this added overhead. As a result, edgeVLM attains a lower total latency than running the SVLM. Moreover, efficiency details are presented in Tab.\ref{tab:exp_efficiency}.

\begin{table}[t]
\centering
\small
\begin{tabular}{c c cc}
\toprule
\multirow{2}{*}{\textbf{System}}
& \multirow{2}{*}{\makecell{\textbf{GPU Memory} \\\textbf{(MiB)}}}
& \multirow{2}{*}{\makecell{\textbf{Prefill} \\\textbf{Time (ms)}}}
& \multirow{2}{*}{\makecell{\textbf{Time /} \\\textbf{Token (ms)}}} \\
\\
\midrule
SVLM Only
  & 10411      & 169.38  & 46.86\\
\midrule
edgeVLM
  & 9589      & 166.81 & 44.75\\
\bottomrule
\end{tabular}
\caption{Efficiency analysis of edgeVLM.}
\label{tab:exp_efficiency}
\vspace{-20pt}
\end{table}

\subsection{Method-level Evaluation}
\vspace{-5pt}
\subsubsection{Ablation Study}
\vspace*{-2pt}
\label{sec:ablation}

We conduct a progressive ablation study on the nuScenes dataset to evaluate the individual and combined contributions of different modules. To isolate component effects, all image frames are uploaded, eliminating the influence of frame selection and upload strategies.

As shown in the Tab.\ref{tab:component_ablation}, the first row corresponds to the baseline where the LVLM and SVLM operate in parallel—using the LVLM output when it is returned in time, and otherwise defaulting to the SVLM. This baseline represents the upper-bound performance of existing cloud–edge collaborative VLM architectures. Compared to this baseline, we incrementally integrate each proposed module. Results show that CRM yields the most significant performance gain. Building on CRM, the integration of the other two methods further enhances overall system performance across all three metrics.

\begin{table}[t]
\centering
\renewcommand{\arraystretch}{1.1}
\setlength{\tabcolsep}{3pt} 
\small
\begin{tabular}{c c c| c c c}
\toprule
\textbf{CRM} & \textbf{GGVR} & \textbf{GRVR}
& \textbf{Micro-F1} & \textbf{Macro-F1} & \textbf{0-1 EM} \\
\hline
\ding{55} & \ding{55} & \ding{55} & 64.36 & 46.55 & 22.35 \\
\ding{51} & \ding{55} & \ding{55} & 70.47 & 55.43 & 28.85 \\
\ding{51} & \ding{51} & \ding{55} & 70.77 & 55.58 & 29.20 \\
\ding{51} & \ding{51} & \ding{51} & \textbf{71.15} & \textbf{57.11} & \textbf{29.41} \\
\bottomrule
\end{tabular}

\caption{Ablation Study. To assess the contribution of each component, we use full-frame upload and LVLM–SVLM parallel inference as the baseline, and evaluate performance improvements as modules are added incrementally.}
\label{tab:component_ablation}
\vspace{-14pt}
\end{table}

\subsubsection{Adaptability Study}
To evaluate the effectiveness of the proposed framework under different combinations of model capacities, we conduct additional experiments by replacing both the SVLMs and LVLMs. All image frames are uploaded as described in Sec.\ref{sec:ablation}. Specifically, we add MiniCPM-o~\cite{yao2024minicpm} and VLM-R1~\cite{shen2025vlm} as the SVLMs and GPT-4o~\cite{hurst2024gpt} as the LVLM. As shown in the Fig.\ref{fig:svlm_generalization}, the proposed edgeVLM coordination strategy consistently improves real-time inference performance across model configurations. VLM-R1 is an SVLM enhanced through distillation and reinforcement learning. Experiments demonstrate that powerful models can be seamlessly integrated into the edgeVLM framework to further enhance system capabilities.

\vspace*{-6pt}
\section{Conclusion}
\label{sec:conclusion}
\vspace{-4pt}
In this work, we propose a novel context transfer paradigm that enables LVLMs to provide real-time guidance to SVLMs through semantic and visual contexts. Building on this paradigm, we introduce edgeVLM, a collaborative framework that integrates two corresponding modules: Context Replacement and Visual Focus. Extensive experiments across diverse tasks, datasets, and model combinations demonstrate that edgeVLM outperforms existing collaboration methods. Moving forward, we envision integrating stronger foundation models and adaptive scheduling strategies into the collaborative framework to further optimize performance in real-world deployments.

{
    \small
    \bibliographystyle{ieeenat_fullname}
    \bibliography{main}
}



\end{document}